\documentclass{article}

\usepackage{arxiv}
\usepackage{mathtools}
\usepackage[utf8]{inputenc} 
\usepackage[mathcal]{eucal}
\usepackage[T1]{fontenc}    
\usepackage{hyperref}       
\usepackage{url}            
\usepackage{booktabs}       
\usepackage{amsfonts}       
\usepackage{nicefrac}       
\usepackage{microtype}      
\usepackage{lipsum}
\usepackage{fancyhdr}       
\usepackage{graphicx}       
\graphicspath{{media/}}     

\usepackage[linesnumbered,ruled,vlined]{algorithm2e} 
\SetKwInput{KwInput}{Input}   
\SetKwInput{KwOutput}{Output}   
\usepackage{multirow}
\usepackage{comment}

\title{Dynamic Adaptive Threshold based Learning for Noisy Annotations Robust Facial Expression Recognition}

\author{
  Darshan Gera \\
  SSSIHL, Brindavan Campus \\
  Bengaluru, Karnataka, India\\
  \texttt{darshangera@sssihl.edu.in} \\
  \And
  Naveen Siva Kumar Badveeti\\
  SSSIHL, Prasanthi Nilayam Campus \\
  Sri Sathya Sai District, Andhra Pradesh, India \\
  \texttt{bnaveensivakumar@gmail.com} \\
  \AND
 Bobbili Veerendra Raj Kumar\\
  SSSIHL, Prasanthi Nilayam Campus \\
  Sri Sathya Sai District, Andhra Pradesh, India \\
  \texttt{veerendra.rajkumar@gmail.com} \\
  \And
  S Balasubramanian \\
  SSSIHL, Prasanthi Nilayam Campus \\
  Sri Sathya Sai District, Andhra Pradesh, India\\
  \texttt{sbalasubramanian@sssihl.edu.in} \\
}

\hypersetup{colorlinks=true, urlcolor=cyan,
pdftitle={Dynamic Adaptive Threshold based Learning for Noisy Annotations Robust Facial Expression Recognition},
pdfauthor={Darshan Gera,  Naveen Siva Kumar Badveeti, Bobbili Veerendra Raj Kumar,S Balasubramanian},
pdfkeywords={Multi Task Learning, Semi Supervised Learning, Facial Expression Recognition, AffWild2, s-aff-wild2},
}

\begin{document}

\maketitle

\begin{abstract}The real-world facial expression recognition (FER) datasets suffer from noisy annotations due to crowd-sourcing, ambiguity in expressions, the subjectivity of annotators and inter-class similarity. However, the recent deep networks have strong capacity to memorize the noisy annotations leading to corrupted feature embedding and poor generalization. To handle noisy annotations, we propose a dynamic FER learning framework (DNFER) in which clean samples are selected based on dynamic class specific threshold during training. Specifically, DNFER is based on supervised training using selected clean samples and unsupervised consistent training using all the samples. During training, the mean posterior class probabilities of each mini-batch is used as dynamic class-specific threshold to select the clean samples for supervised training. This threshold is independent of noise rate and does not need any clean data unlike other methods. In addition, to learn from all samples, the posterior distributions between  weakly-augmented image and  strongly-augmented image are aligned using an unsupervised consistency loss. We demonstrate the robustness of DNFER on both synthetic as well as on real noisy annotated FER datasets like RAFDB, FERPlus, SFEW and AffectNet. Our source codes are publicly available at \url{https://github.com/1980x/DNFER}.
\end{abstract}



\maketitle

\section{Introduction}
\begin{figure}[hbt!]
\centerline{\includegraphics[width=1.5in]{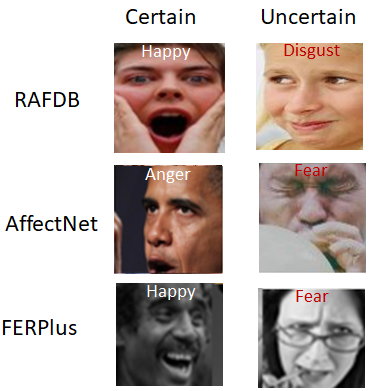}}
\caption{Examples of samples with certain and uncertain annotations}
\label{sample}
\end{figure}

\begin{figure}[hbt!]
\centerline{\includegraphics[width=3in, height=2in]{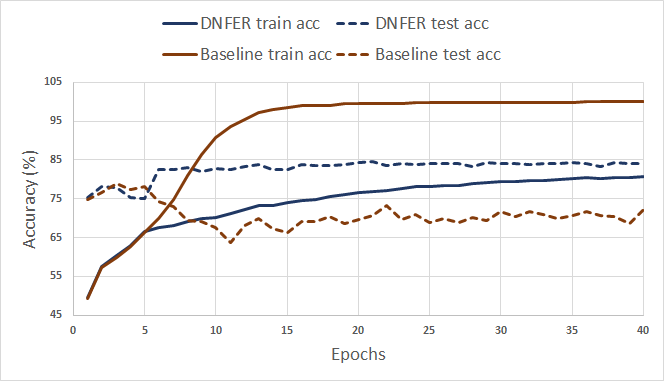}}
\caption{Performance comparison of train and test set accuracy on RAFDB with 30\% synthetic label noise with Baseline (supervised training) and DNFER (proposed) }
\label{overfitting_rafdb}
\end{figure}

Facial expressions recognition (FER) is aimed at making machines understand human emotions and intentions. In the recent years, a significant progress is made towards developing deep learning (DL) based robust FER systems \cite{oadn, scan, LNLAttenNet, ran, FDRL, EfficientFace} along with development of large-scale datasets like RAFDB \cite{rafdba,rafdbb}, FERPlus \cite{ferplus}, AffectNet \cite{affectnet}, SFEW \cite{sfew} etc. However, large real-world datasets suffer from noisy annotations due to subjectivity of annotators, ambiguity of different expressions, poor quality of images etc. A sample of images from RAFDB, FERPlus and AffectNet are shown in Fig. \ref{sample}. Clearly, some of them have reliable labels whereas others have uncertain labels. These uncertain labels lead to noisy annotations leading to corrupted features and poor generalization as shown in the Fig. \ref{overfitting_rafdb}. Here, we train ResNet-18 \cite{resnet} on RAFDB with standard cross-entropy loss (denoted as Baseline) in the presence of 30\% synthetic label noise and compare the performance on train and test sets during training. As training progress, Baseline model overfits noisy samples (shown in solid brown color) but its performance on test set (shown in dotted brown color) is much lower compared to the proposed method (shown in blue color). This is due to strong memorization capability of deep networks \cite{memorize_1, memorize_2}, as network overfits on noisy annotated samples. 

\begin{figure*}[hbt!]
\centerline{\includegraphics[width=5.2in]{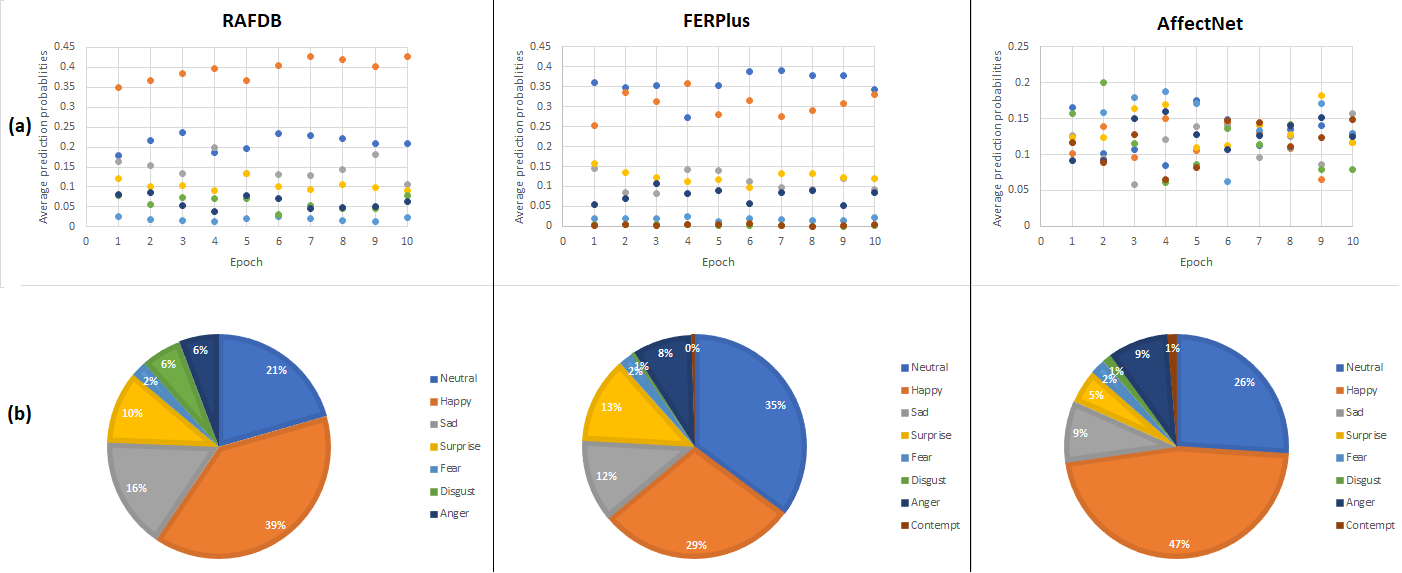}}
\caption{(a) The average posterior prediction probabilities of different expression classes during the first ten epoch of training for RAFDB, FERPlus and AffectNet datasets, (b) The sample size imbalance of different expression classes for benchmark FER datasets}
\label{fig_motivation}
\end{figure*}

Recently, some works like \cite{scn, gera_ambigious_annotations, ulcag, ldl, DMUE, gera_cct1} have studied the problem of uncertain labels in FER but their performance is not satisfactory. Self-cure network (SCN) \cite{scn} attempts to eliminate the influence of noisy annotated samples by loss re-weighing and relabelling incorrect ones based on learnt importance scores but it suffers from self-confirmation bias. ULC-AG \cite{ulcag} improvises over SCN by using auxiliary action unit (AU) graphs for uncertain label correction. LDL \cite{ldl} learns the label distribution whereas DMUE mines the latent distribution and uses pairwise uncertainty estimation for ambiguity estimation. In \cite{gera_ambigious_annotations}, likelihood of sample being clean is obtained based on Jensen Shannon Divergence between prediction probabilities and soft ground-truth labels. Noise robust loss function called Rayleigh loss \cite{rayleigh_loss} is proposed to learn discriminative expression features in the presence of noisy annotations. However,  many of these methods fail to perform well and most of these are tested only on synthetic noisy datasets. Robust training under noisy labels is an important problem in DL. Recently, many approaches have been developed like robust loss functions \cite{loss_imae, loss_symmetric}, loss correction \cite{loss_correction, noise_adaptation} and sample selection methods \cite{mentornet, co-teaching, co-teaching_plus, jocr, nct}, etc. Sample selection methods like Co-teaching \cite{co-teaching}, Co-teaching+ \cite{co-teaching_plus}, JoCoR \cite{jocr} and NCT \cite{nct} train two networks with peer learning or joint learning using selected clean samples based on small loss criterion but they need to know the noise rate to select small loss samples. Further, threshold chosen to select clean samples is same for all the classes and fixed throughout of training which may not be true. This is demonstrated in the Fig. \ref{fig_motivation} (a). The average posterior prediction probabilities of different expression classes Happy, Neutral, Sad, Fear, Disgust, Surprise, Anger and Contempt for benchmark FER datasets RAFDB (except contempt), FERPlus and AffectNet are plotted during the first 10 epochs of supervised training. These average class probabilities are different for each class as well as vary while training. In addition, the number of samples are imbalanced in terms of different expression classes for different datasets as shown in Fig. \ref{fig_motivation} (b). So, fixing a same prior threshold for all classes is likely to select samples only from majority classes which leads to poor performance on other minority classes. Further, having a same threshold for each mini-batch throughout training prevents selection of hard samples. Motivated by this observation, we propose a dynamic adaptive threshold for selecting clean (reliable) samples based on average posterior prediction probabilities of samples in each mini-batch. The selected clean samples are trained using supervision loss based on cross-entropy (CE) loss . Further, in order to make use of all the samples for feature learning, the posterior distributions of weakly-augmented image and strongly-augmented image of each sample is minimized using unsupervised consistency loss based on symmetric Kl-divergence loss. The proposed framework uses a single shared network for noise robust FER learning based on dynamic adaptive threshold named as DNFER. The proposed DNFER framework is illustrated in Fig. \ref{framework} and explained in Section \ref{Section_III}. Effectiveness of the DNFER framework is validated on synthetic as well as real noisy annotated FER datasets. 

Overall, the main contributions can be summarized as follows:
\begin{enumerate}
    \item The proposed DNFER selects clean (reliable) samples for supervised learning based on dynamic class adaptive threshold and utilizes noisy (unreliable) samples using consistency loss for facial expression representation learning.
    \item DNFER is architecture independent without depending upon noise rate as well as separate clean data. It uses a data-dependent class-specific dynamic threshold to select clean samples in each mini-batch.
    \item Our DNFER is an end-to-end framework and achieves superior performance on several benchmark FER datasets as well as on synthetic and real noisy annotated datasets.
    
\end{enumerate}

The rest of the paper is organized as follows: Section \ref{Section_II} reviews the related work in the area of FER in general and with noisy label learning as well as general DL based noisy robust methods. Section \ref{Section_III} presents the proposed DNFER framework. The effectiveness of DNFER is demonstrated by comparative performance evaluation and ablation study in Section \ref{Section_IV}. Finally, Section \ref{Section_V} concludes the work.

\section{Related Work}\label{Section_II}
In this section, we first briefly review the related work in FER with and without noisy labels followed by noisy label learning.

\subsection{Facial Expression Recognition}
Recent DL based FER methods are aimed at either representation learning \cite{oadn, scan, LNLAttenNet, ran, FDRL, EfficientFace} or uncertain label based learning \cite{RUL,DMUE, ldl, scn, ulcag}. Attention based CNN methods like SCAN \cite{scan}, OADN \cite{oadn}, GACNN \cite{gacnn}, RAN \cite{ran} learn local and global features based on attention scores of different facial regions for learning robust expression features in the presence of occlusion and pose variations. FDRN \cite{FDRL} uses feature decomposition and reconstruction networks respectively for capturing similarities across different expressions and distinctive feature for each expression. Efficient-face \cite{EfficientFace} is a light-weight model designed using dept-wise convolution and channel modulator for capturing local-global features for robustness against occlusion and pose variations. In the same line, CERN \cite{cern} is a low-calorie power-packed architecture with 1.45M parameters which is robust to occlusion and pose variations. Although, DL based methods have achieved great success for robustness against occlusion and pose variations in FER, their unique and strong memorization learning ability makes them overfit on the noisy annotations \cite{memorize_1, memorize_2}. 

Recently, some works like SCN \cite{scn}, ULC-AG \cite{ulcag}, DMUE \cite{DMUE}, LDL \cite{ldl}, IPA2LT \cite{ipa2lt}, CCT \cite{gera_cct1} and RUL \cite{RUL} have attempted to mine the latent label distribution to correct the uncertain labels in FER datasets. SCN \cite{scn} learns the confidence score of each sample for loss-reweighing using a self-attention module. It corrects the labels of uncertain samples (low score samples) based on prediction probability. In Inconsistent Pseudo Annotations to Latent Truth (IPA2LT), multiple inconsistent human and machine predicted labels are assigned to each sample, and subsequently the true label is learnt by maximizing the log-likelihood of these inconsistent annotations. Rayleigh loss \cite{rayleigh_loss} introduces a constraint to ensure that class center is far away from nearest neighbours along with center loss. RUL \cite{RUL} utilises a model to learn uncertainties of input images and uses these uncertainties to mix the features of two different labels. In CCT \cite{gera_cct1}, three networks are trained using a convex combination of supervision loss and consistency combined using a dynamic balancing scheme to combat the influence of noisy annotated samples. LDL \cite{ldl} proposed a label distribution learning using auxiliary-label space graphs based on action-units similarity with  different expressions. DMUE \cite{DMUE} designed an uncertainty estimation module which dynamically learns from mined label distribution and ground truth label of the samples. Though these methods mine the label distribution but these have not observed effect of noisy labels in the presence of inter-class sample size imbalance and intra-class imbalance of difficulty level. To mitigate this, a dynamic class adaptive learning based framework is proposed for robust expression representation learning in the presence of noisy annotated samples .
\subsection{Noisy Labels Robust Deep Learning} Learning with noisy labels is an important research problem in DL. Recently, it has been observed that DNNs first fit clean data (data with correct labels), and then, towards later part of the training, memorize examples with noisy annotations \cite{memorize_1, memorize_2}. This makes it challenging to train models efficiently on real world datasets in which poor quality annotations are obtained using online queries or crowd sourcing. Based on above, many methods have been proposed which can be categorized as sample selection using small loss \cite{mentornet, co-teaching, co-teaching_plus}, label cleansing \cite{label_cleaning_iterative}, weighting \cite{reweight}, robust loss functions \cite{loss_imae, loss_symmetric} etc. Please refer to survey \cite{survey_noisy} for detail on these methods. This paper adopts the strategy of sample selection, so we briefly review these here. These methods rely on selecting small loss samples with or without co-training \cite{decoupling, co-teaching_plus, jocr, co-teaching, nct} due to the fact that small loss samples are associated with clean labels. Inspired from semi-supervised learning consisting of labelled and unlabelled data, Co-training \cite{cotraining} principle has been followed in these works. Co-training assumes that each data has two different views and each view is sufficient for learning an effective model. These views can either correspond to different data sources or different representations, and provide complementary information about each sample. Co-training assumes that models trained on two different views have consistent predictions on unlabelled data. Co-training based methods follow the principle of joint training with consensus in JoCoR [139] or mutual training with consensus in NCT \cite{nct} or peer training with cross supervision in Co-teaching \cite{co-teaching} and Co-teaching+ \cite{co-teaching_plus}. Co-teaching and JoCoR follow the principle of agreement whereas Decoupling and Co-teachng+ depend upon the “Disagreement” strategy to keep the two networks diverged to achieve better ensemble effects. To summarize, recent methods co-train two networks using small loss samples based on agreement or disagreement combined with some kind of regularization. These methods need to know noise rate to drop high loss samples in each mini-batch which may be difficult to estimate in real world setting. Further, threshold chosen based on noise rate to select clean samples is same for all the classes and fixed throughout training which may not be true. Recent works like CCT \cite{gera_cct1} and Co-curing \cite{geracocuring} have applied these methods in FER scenario but these use multiple networks for training which leads to computational burden. Further, they use a fixed threshold for all expression classes but inherently FER datasets are imbalanced in sample size as well as intra-class samples have serious imbalance in the difficulty level. This imbalance factor has not been considered for selecting samples of small loss during learning.

\section{Proposed Method}\label{Section_III}
\subsection{DNFER Overview}
\begin{figure*}[hbt!]
\centerline{\includegraphics[width=5in]{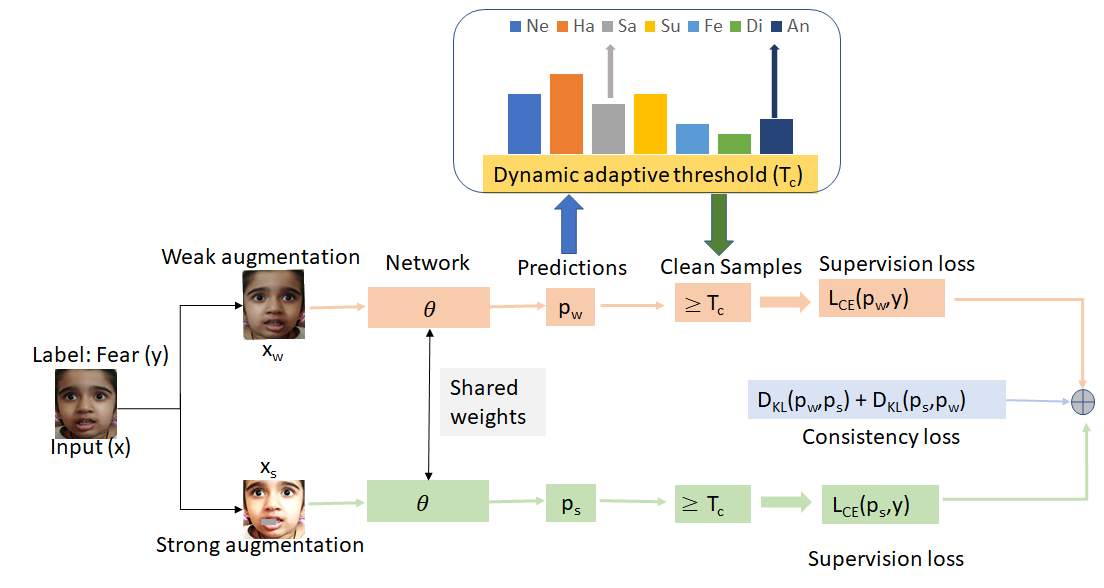}}
\caption{Proposed DNFER framework: a) Given an input face image $x$ with expression label $y$, its weak-augmented image ($x_{w}$) and strongly-augmented image ($x_{s}$) are input to a shared network $f$ with parameter $\theta$. Clean samples are selected based on dynamic class specific threshold ($T_{c}$) computed using average posterior probabilities of samples in each mini-batch. Model is trained using supervision loss based on selected clean (reliable) samples and unsupervised consistency loss by aligning the posterior predictions $p_{w}$ and $p_{s}$ of all samples.  }
\label{framework}
\end{figure*}

The proposed DNFER method is based on the idea of small loss (reliable) sample selection \cite{memorize_1, memorize_2} using a dynamic adaptive threshold and uses 2 different transformed (augmented) views of an image based on co-training \cite{cotraining} for learning noisy label robust expression feature representation and classification. Unlike the previous works \cite{co-teaching, co-teaching_plus, jocr}, which use a fixed threshold for all classes for selection of small loss samples depending upon the noise rate in the data (which may not be known and otherwise difficult to estimate), our method dynamically learns the adaptive threshold based on the average posterior probabilities of each mini-batch without depending upon noise rate in the given data. Since posterior probabilities are inversely related to loss values in the categorical CE loss for supervision, so, small loss samples (high posterior probabilities) are likely to correspond to clean (reliable labels). DNFER uses a dynamic class adaptive threshold which improves along with learning taking into account intra class sample difficulty as well as inter class sample size imbalance. The selected clean samples are trained using CE loss. Further, unreliable samples are used for expression feature representation by aligning the posterior distribution of weak-augmented image and strongly-augmented image using unsupervised consistency loss. The supervision loss and consistency loss are combined using a weighing parameter (chosen based on ablation study). The overview of the proposed DNFER framework is shown in Fig.\ref{framework}.

\subsection{Problem formulation}
Given a batch of N images $S =\{(x_{i}, y_{i})\}_{i=1}^{N}$ where each facial image $x_{i}$ has an  expression label $y_{i}\in\{1,2,...,C\}$ and C denotes the number of expression classes. The shared backbone network is parameterized by $\theta$ (in this work ResNet-18 is used as the backbone). Features from backbone are classified using a fully connected layer (FC) layer with softmax to obtain prediction probabilities. The weak-augmented image and strongly-augmented image are denoted as $x_w$ and $x_s$ and their prediction probabilities as $p_w$ and $p_s$ respectively. Standard cropping along with horizontal flipping are used for weak augmentations. Randaugment \cite{randaugment} is used as strong augmentation. During training, the small loss samples from  each mini-batch $S_n$ are selected based on dynamic adaptive confidence threshold $T_c$ (explained below). The model is trained by minimizing a loss $L$ which is a convex combination of supervision loss (based on standard CE loss) denoted as $L_{sup}$ and consistency loss (based on symmetric KL divergence) denoted as $L_{cons}$. The overall loss is given as below :

\begin{equation} \label{loss}
L = \alpha * L_{cons}  + (1 - \alpha) * L_{sup} 
\end{equation}
Here,  $\alpha \in [0,1]$. 

\subsection{Dynamic Adaptive Threshold based Clean Sample Selection}
As deep networks tend to fit easy (small loss) samples before the noisy (high loss) ones, we use CE loss for training the model using ground truth labels, denoted as $L_{CE}((X,Y);\theta)$,
\begin{equation} \label{CELOSS}
    L_{CE} = {\{l_{i}\}}_{i=1}^{N} = \{-\sum_{c=1}^{C} y_{i=1}^{c} \log( p^{c}(x_{i};\theta)\}_{i=1}^{N}
\end{equation}

where $x_{i} \in X$, $y_{i} \in Y$, $p^{c}$ is observed posterior probability for class $c \in \{1,2,...,C\}$ from network's softmax layer. Clean samples are distinguished from noisy ones according to $L_{CE}$. Since, the loss and posterior probabilities are inversely related, we use class wise mean of posterior probabilities of samples in a mini-batch to obtain class wise threshold $T_{c}$. It assumes that all sample losses are independent and identically distributed. Samples with predicted probabilities greater than the dynamic threshold $T_{c}$ are treated as clean samples. $T_{c}$  is computed as :

\begin{equation} \label{dynamicthreshold}
T_{c} = \frac{1}{\lvert S_{c} \rvert}  \sum_{x \in S_{c}} p^{c}(x;\theta) 
\end{equation}

where $S_{c}$ is set of samples in the current mini-batch with class c.
This dynamic class wise adaptive threshold evolves along with learning depending upon the mini-batch and takes into account the intra-class difficulty of image from same expression class as well as inter-class imbalance of sample sizes for different expression classes. As shown in Fig \ref{fig_motivation}, happiness, neutral, surprise and sad expression classes are easily recognizable compared to fear, disgust and contempt in real world FER datasets \cite{rafdba, ferplus, affectnet}. Due to which, these datasets are imbalanced in terms of sample size. Apart from this, within one expression class, some images are easy and others are hard apart from noisy annotated samples. As deep networks fit easy samples before hard and noisy, so network is trained using CE loss on easy samples using ground truth labels for warm up epochs (hyper-parameter based on ablation study) which results in different loss distribution for different expression classes. Average of class probabilities are different for different expression classes as shown in the Fig. \ref{fig_motivation} (a) and these change as training evolves. So, this dynamic class specific adaptive threshold adopts intra-class and inter-class imbalance for selecting clean samples. Finally, based on dynamic threshold $T_{c}$, clean-sample selection is done as follows:
\begin{equation} \label{cleansampleselection}
S_{c}^{clean} =  \{ x \in S_{c} \ni p^{c} \geq  T_{c} \}
\end{equation}

\subsection{Supervision loss} After the warm-up epochs, network is trained based on supervision loss using selected clean samples. Inspired by Co-training methods \cite{cotraining, co-teaching, jocr} for semi-supervised learning and noisy label training, supervision loss is computed based on selected clean samples for two different transformed views of input samples in each mini-batch. The posterior probabilities of selected clean weak-augmented ($x_{w}^{clean}$) and strongly-augmented ($x_{s}^{clean}$) transformed images are denoted as $p_{w}^{clean}$  and $p_{s}^{clean}$ respectively.
\begin{equation} \label{sup_loss}
 L_{sup} = L_{CE}(y^{clean},p_{w}^{clean})  + L_{CE}(y^{clean},p_{s}^{clean})
\end{equation}
where $y^{clean}$ is the ground-truth labels of selected clean samples ($S_{c}^{clean}$) in current mini-batch.

\subsection{Consistency loss} Finally, in order to utilize all the samples for learning, the posterior distributions of weak-augmented and strongly-augmented images are aligned using unsupervised consistency loss according to Eq. \eqref{cons_loss}. The symmetric Kullback-Leibler divergence ($D_{KL}$) defined in Eq. \eqref{KLLoss} is used to compute consistency loss.

\begin{equation} \label{cons_loss}
 L_{cons}  =  D_{KL}(p_{s}||p_{w}) + D_{KL}(p_{w}||p_{s})
\end{equation}

\begin{equation} \label{KLLoss}
 D_{KL}(p_{k}||p_{j}) =  \sum_{l=1}^{C} p_{k}^{l} \log {\frac{p_{k}^{l}}{p_{j}^{l}}}
\end{equation}

\subsection{DNFER training algorithm}
 During the warm-up period, all the samples are used for training based on supervision loss. After warm-up, model is trained using $L_{sup}$ and $L_{cons}$. Details of training are provided in Algorithm \ref{Training_algorithm}. The loss weighing factor $\alpha$ to combine $L_{sup}$ and $L_{cons}$ in Eq. \eqref{loss} is fixed as 0.5 (here warm-up period as well as loss weighing parameter are chosen based on ablation study).

\begin{algorithm}
\DontPrintSemicolon
  \KwInput{ Given a model f with parameters $\theta$, dataset $S =\{(x_{i}, y_{i})\}_{i=1}^{N}$ , mini-batch size (b), learning rate($\eta$),  number of expression classes (C), total epochs $E_{max}$, warm-up epochs ($E_{warm}$)}
  \KwOut{Updated model parameters $\theta$}
   Initialize $\theta$ randomly.\;
  
   \For{$e = 1,2,..,E_{max}$} 
   {
    Shuffle training samples $\{(x_{i}, y_{i})\}_{i=1}^{N}$\;
    Sample mini-batch $S_{n}$ from S\;
    \For{ each class $c \in \{ 1,2,..,C\}$} 
    {
    Compute dynamic confidence threshold $T_{c}$ using Eq. \eqref{dynamicthreshold}\;
    Select clean samples $S_{c}^{clean}$ from current mini-batch using Eq. \eqref{cleansampleselection}\;
    }
    Compute supervision loss $L_{sup}$ on above selected clean samples using Eq. \eqref{sup_loss} \;
    Compute consistency loss $L_{cons}$ on all samples using Eq. \eqref{cons_loss} \;
    Compute total loss $L$ using Eq. \eqref{loss} \;
     \uIf{ $e < E_{warm}$ }{
     $\alpha = 0$ \;
  }
  
  \Else{
    $\alpha = 0.5$ \;
  }
    Update $\theta= \theta - \eta \nabla L_{\theta}$ \;
     }   
   return $\theta$
\caption{DNFER training algorithm}
\label{Training_algorithm}
\end{algorithm}

\section{Experiments}\label{Section_IV}
The proposed DNFER is compared with recent state-of-the-art (SOTA) methods on popular real-world benchmarks as well as in the presence of synthetic label noise on FER datasets. Further, we conduct extensive experiments to verify the component wise contribution of each part of DNFER. Ablation studies are conducted for hyperparameter selection. Experimental results demonstrate that our DNFER is robust in presence of noisy annotations present in FER datasets. First, we briefly introduce the experimental setup and next present the performance evaluation followed by ablation studies and comparison with SOTA methods.
\subsection{Experimental Setup}
\subsubsection{Datasets}
We evaluate DNFER on four commonly used datasets RAFDB, FERPlus, SFEW and AffectNet.

\begin{itemize}

   \item \textbf{RAFDB} \cite{rafdba, rafdbb}: The Real-world Affective Face Database (RAFDB) is constructed using 29762 face images with two different subsets: i) basic emotion subset annotated with 7 emotions (i.e. happy, sad, neutral, surprise, disgust, angry, fear), ii) compound emotion subset. In this study, we use basic emotion set of 12271 images for training and 3068 images for testing. Both train and test sets are imbalanced w.r.t sample size of different expression classes.
   
   \item 	\textbf{FERPlus} \cite{ferplus}: FERPlus is an extended version of FER2013 \cite{fer2013}. It consists of images with the 8-basic emotions (with contempt), of which 28709 are used for training, 3589 are used for validation, and the remaining 3589 for test. As 10 independent annotators are used for labeling each image, label quality of FERPlus is superior compared to FER2013. For a fair comparison, majority voting is used for reporting performance.
   
   \item \textbf{SFEW} \cite{sfew}: SFEW contains images from movies with seven basic emotions with train set of 891 samples, validation set of 431 samples and test set of 372 samples. Performance is reported on the validation set as test sets labels are not released.
   
   \item \textbf{AffectNet} \cite{affectnet}: AffectNet is the largest dataset with 0.44M manually annotated and 0.46M automatically annotated facial expression images for 8 emotions. Main limitation is that each image has been labelled with a single annotator due to which labels are not reliable. AffectNet-7 refers to manually annotated set without contempt class whereas AffectNet-8 includes all expression images. AffectNet-7 has 283901 images for training and 3500 images for validation whereas AffectNet-8 has 287568 images for training and 4000 images for validation. Test set labels are not released, so, being consistent with previous works, validation sets are used for reporting results. In addition, we use \textbf{Automatically annotated subset} for training under real noisy conditions.
   
   \item \textbf{Synthetic noisy annotated datasets}: We randomly change 10\%, 20\% and 30\% labels of training images from RAFDB, FERPlus and AffectNet-8 datasets to create synthetic noisy annotated datasets. Performance of DNFER is reported on the corresponding clean test/validation sets.
   
\end{itemize} 
   
\subsubsection{Performance Metrics} We use overall test accuracy (\%) to report the performance for all algorithms. Beside we conduct experiments three times using different random seeds to obtain the mean accuracy and their standard deviations. Confusion plots are provided to obtain individual expression performance.

\subsubsection{Implementation Details} In the below experiments, we use MTCNN \cite{mtcnn} to detect and resize the facial expression images with size of 224x224. Our proposed DNFER framework is implemented in Pytorch DL toolbox using single GeForce GTX TitanX GPU with 12GB RAM. Backnone network used is ResNet-18 which is pretrained on MS-Celeb-1M \cite{msceleb} face recongition dataset similar to previous works. Random horizontal flipping with probability 0.5 is used for weak augmentation beside random crop with 4 pixel and resize to 224x224. For strong augmentation, RandAugment \cite{randaugment} is used. RandAugment randomly select two transformations from a group of transformations like contrast adjustment, rotation, color inversion, translation, etc. Oversampling\footnote{https://github.com/ufoym/imbalanced-dataset-sampler} is used to overcome class imbalance problem in AffectNet similar to \cite{ran,scn, oadn}. During training, the batch size used is 128, the max epochs (Emax) is set as 40 for RAFDB, FERPLus datasets and 20 for AffectNet dataset, warm up period is chosen as 5 for RAFDB, FERPlus datasets and 3 for AffectNet dataset. Model is optimized using Adam optimizer with an initial learning rate of 0.001 which is decayed exponentially with a factor of 0.95 every epoch. 

\subsection{Experiment Results on Synthetic Noisy Annotated Datasets}

\begin{table}[hbt!]
\caption{Performance evaluation (accuracy \%) on FER datasets with synthetic label noise }
\label{table}
\small
\centering
\setlength{\tabcolsep}{3pt}
\resizebox{0.7\textwidth}{!}{
\begin{tabular}{c|c|c|c|c}
\hline
Noise-level  &   Method    &  RAFDB                     & FERPlus        & AffectNet-8  \\
  \hline
  \multirow{6}{*}{10\%}     & Baseline    &  81.74      &  85.87          & 57.21                  \\
                            & SCN \cite{scn}        &  82.18        &  84.28          & 58.58                  \\
                            & RR  \cite{rayleigh_loss}       &  82.43        &  83.93          & 60.04                  \\
                            & ULC-AG \cite{ulcag}     & 83.21        &  -         & 59.45  \\
                            & DMUE  \cite{DMUE}     & 83.19$\pm$ 0.83  & -           & 61.21$\pm$0.36 \\
                            \cline{2-5}
                            & \textbf{DNFER}      & \textbf{88.09$\pm$0.37}        &  \textbf{88.67$\pm$0.22}          & \textbf{59.94$\pm$0.08}  \\
\hline
 \multirow{6}{*}{20\%}      & Baseline    & 79.60        & 84.02           & 56.23                   \\
                            & SCN         & 80.01        &  83.17          & 57.25                   \\
                            & RR        &  80.41        &  83.55          & 58.47                  \\
                            & ULC-AG     & 81.16        &  -         & 58.51  \\
                            & DMUE       & 81.02$\pm$ 0.69  & -           & 59.06$\pm$0.34 \\
                            \cline{2-5}
                            & \textbf{DNFER}      & \textbf{86.94$\pm$0.17}        &  \textbf{87.95$\pm$0.26}          & \textbf{58.63$\pm$0.18}  \\
 \hline             
 \multirow{6}{*}{30\%}      & Baseline    & 74.31        & 82.30           & 52.60                   \\
                            & SCN         & 77.46        & 82.47           & 55.05                   \\
                            & RR        &  76.77        &  82.75          & 55.82                  \\
                            & ULC-AG     & 79.01        &  -         & 56.45  \\
                            & DMUE       & 79.41$\pm$ 0.74  & -           & 56.88$\pm$0.56 \\
                            \cline{2-5}
                            & \textbf{DNFER}      & \textbf{84.99$\pm$0.35}        &  \textbf{86.58$ \pm$0.16}          & \textbf{56.55$\pm$0.32}  \\

\hline             
\end{tabular}
}

\label{table_dnfer_tab_syntheticnoise}
\end{table}

Synthetic symmetric noise is manually added on RAFDB, FERPlus and AffectNet-8 datasets by randomly changing labels in the ratio of 10-30\%. We compare  DNFER with single ResNet-18 trained using CE loss (referred as Baseline) and against recent noise robust SOTA FER methods like SCN \cite{scn}, RR \cite{rayleigh_loss}, ULC-AG and DMUE. Results are presented in Table \ref{table_dnfer_tab_syntheticnoise}.  Clearly, their performance is much inferior compared to our proposed DNFER in the presence of different noise levels on RAFDB and FERPlus (ULC-AG and DMUE don't report on FERPlus). There is gain of around 5\% for different noise levels on RAFDB compared to next best performing DMUE. DNFER performs best on FERPlus compared to Baseline as well as SCN and Rayleigh loss (denoted as RR). AffectNet-8 being a challenging noisy dataset because of heavy class imbalance as well as intra class difficulty, no method is able to achieve good performance. DNFER obtains comparable performance to that DMUE on AffectNet-8 in the presence of noisy labels.

\subsection{Evaluation on Real Noisy Annotated Dataset}
Unlike synthetic label noise, real-noisy annotated FER datasets can include label noise of unknown level. We train DNFER using automatically annotated subset of AffectNet-7 and test on corresponding validation set. We compared the performance w.r.t recent SOTA FER methods like RAN, GACNN and OADN and noise robust FER methods like SCN and ULC-AG. As shown in Table \ref{table_realnoise}, DNFER performs best in the presence of real uncertainties compared to general as well as noisy label robust FER methods ULC-AG and SCN. Uncertainties in real datasets are due to imbalance of sample size in different classes as well as imbalance of intra class sample at difficulty level. Due to adaptive threshold for different classes dynamically evolving along with learning, DNFER mitigates these real ambiguities effectively compared to recent methods. 
\begin{table}[hbt!]
\caption{Performance evaluation on real noisy annotated AffectNet-7 (* denotes FER methods with noisy labels robust learning)} 
\centering
\small
\setlength{\tabcolsep}{3pt}
\begin{tabular}{c|c|c}
                            \hline
                            Method       & Year  & AffectNet-7    \\
                            \hline
                             Baseline    &  -    &  53.85                            \\
                             RAN \cite{ran}         &  2019 &  56.43                           \\
                             GACNN \cite{gacnn}      &  2019 &  52.43                           \\
                             OADN  \cite{oadn}      &  2020 &  55.37                            \\
                             SCN*    \cite{scn}     &  2020 &  55.43                           \\
                             ULC-AG* \cite{ulcag}      &  2022 &  57.37 \\
                             \hline
                            DNFER          &  -   &  \textbf{57.65}           \\
                             \hline
\end{tabular}
\label{table_realnoise}
\end{table}

\subsection{Ablation Study}
\subsubsection{Evaluation of components}
\begin{figure}[hbt!]
\centerline{\includegraphics[width=3.5in]{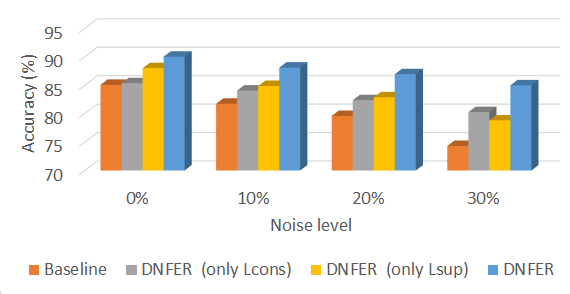}}
\caption{Evaluation of the components in DNFER on RAFDB for different noise levels of 0-30\%}
\label{component_evalaution_RAFDB}
\end{figure}

\begin{figure}[hbt!]
\centerline{\includegraphics[width=3.5in]{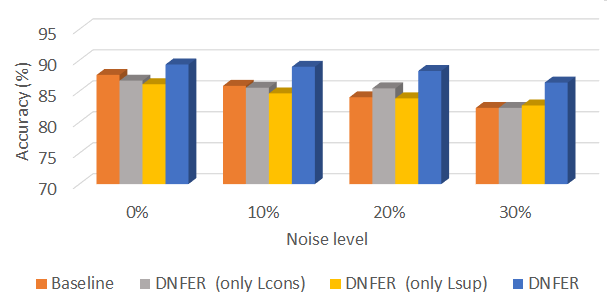}}
\caption{Evaluation of the components in DNFER on FERPlus for different noise levels of 0-30\%}
\label{component_evalaution_FERPLUS}
\end{figure}

DNFER is aimed at the influence of noisy annotated samples for supervision learning based on selected clean samples using dynamic adaptive threshold and making use of all samples for feature representation learning using consistency loss. In this experiment, we design 4 settings for validating the effectiveness of the proposed DNFER. In the first setting, model is trained using all samples based on CE loss for all 40 epochs (referred to as Baseline). In the second setting, DNFER (only $L_{cons}$), model is trained with CE loss for warm-up epochs and later using only $L_{cons}$ i.e. without $L_{sup}$ loss on selected clean samples. In the third setting, DNFER (only $L_{sup}$), model is trained with CE loss for warm-up epochs and later with only $L_{sup}$ loss on selected clean samples without $L_{cons}$. Last one refers to our proposed DNFER. We evaluate the performance in the presence of 0-30\% on RAFDB in the Fig. \ref{component_evalaution_RAFDB} and on FERPlus in the Fig. \ref{component_evalaution_FERPLUS}. Baseline performance on RAFDB (0\% label noise) is inferior compared to DNFER with $L_{cons}$ and DNFER with $L_{sup}$ whereas it is not same on FERPlus (0\% label noise). This indicates that FERPlus has cleaner annotations compared to RAFDB. As noise labels increase from 10-30\% on RAFDB and FERPlus, both DNFER with $L_{cons}$ and DNFER with $L_{sup}$ fail to perform. DNFER using both the components obtains superior performance by selecting clean annotated samples for supervised training and effectively learn the expression features using all the samples in the presence of noisy annotations.

\subsubsection{Impact of Weighting Parameter}
\begin{figure}[hbt!]
\centerline{\includegraphics[width=3.5in]{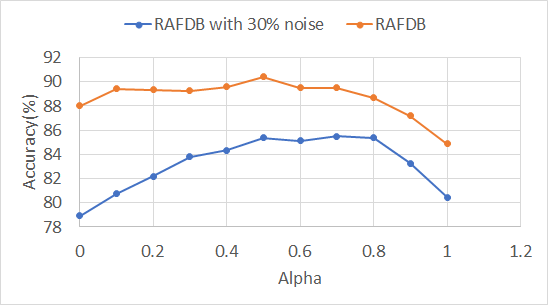}}
\caption{Impact of weighing parameter $\alpha$ on RAFDB (with and without noise)}
\label{weighingparameter}
\end{figure}

We analyse the impact of loss weighing parameter $\alpha$ in Eq. \eqref{loss} to combine the supervision and consistency loss. Fig. \ref{weighingparameter} depicts the performance against various value of $\alpha$ on RAFDB with 30 \% (shown in blue color) and 0\% (shown in orange color) synthetic noise. Clearly, $\alpha=0.5$ gives best performance.
\subsubsection{Influence of Warm-up Epochs}
\begin{figure}[hbt!]
\centerline{\includegraphics[width=3.5in]{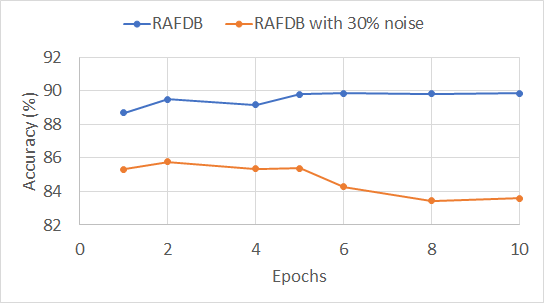}}
\caption{Influence of different number of warm-up epochs on RAFDB (with and without noise)}
\label{warmupepochs}
\end{figure}

In order to analyse the influence of different number of warm-up epochs, we compare the performance on RAFDB with 30 \% (shown in orange color) and 0\% (shown in blue color) synthetic noise as shown in the Fig. \ref{warmupepochs}. As deep networks learn from easy (small loss) samples during initial stage of training but tend to overfit on noisy annotated samples as training progress, we choose warm-up epochs as 5 for reporting results on RAFDB and FERPlus datasets and 3 on AffectNet.
\subsubsection{Comparison of noise over-fitting}

\begin{figure}[hbt!]
\centerline{\includegraphics[width=3.5in]{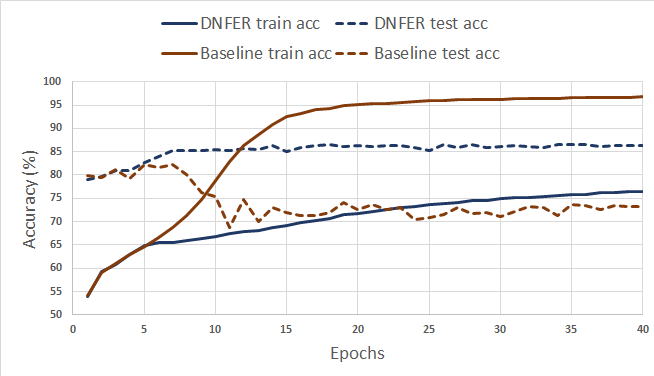}}
\caption{Noise overfitting on FERPlus with 30\% synthetic label noise}
\label{overfitting_ferplus}
\end{figure}

\begin{figure}[hbt!]
\centerline{\includegraphics[width=3.5in]{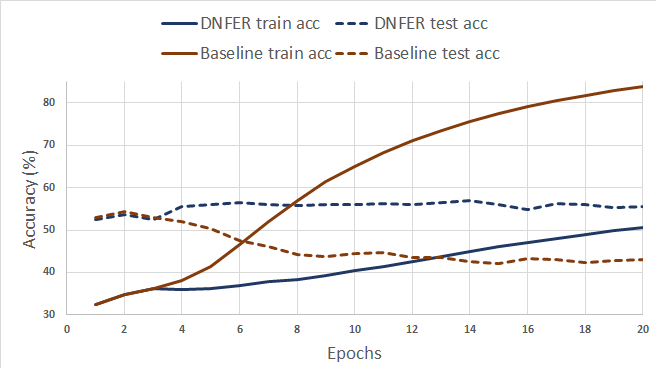}}
\caption{Noise overfitting on AffectNet8 with 30\% synthetic label noise}
\label{overfitting_affectnet}
\end{figure}

In order to analyse the influence of DNFER in mitigating overfitting on noisy annotated samples, we plot the performance on train and test set in the presence of 30\% synthetic label noise on RAFDB (shown in Fig. \ref{overfitting_rafdb}), FERPlus (shown in Fig. \ref{overfitting_ferplus}) and AffectNet (shown in Fig. \ref{overfitting_affectnet}) datasets. It is clear that using Baseline model based on all samples, training accuracy (shown in sold grey curve) saturates whereas test accuracy (shown in dotted curve grey) initially increases but then drops off as training progress. In contrast, DNFER training accuracy (shown in solid blue curve) is much lower compared to Baseline train accuracy and test accuracy curve (shown in dotted blue) is much above to that of Baseline test accuracy. This behaviour is observed across all the three datasets demonstrating the validity of DNFER in mitigating overfitting on noisy annotated samples.

\subsection{Visualization}
\subsubsection{Confusion Plots}
\begin{figure*}[hbt!]
\centerline{\includegraphics[width=5.2in]{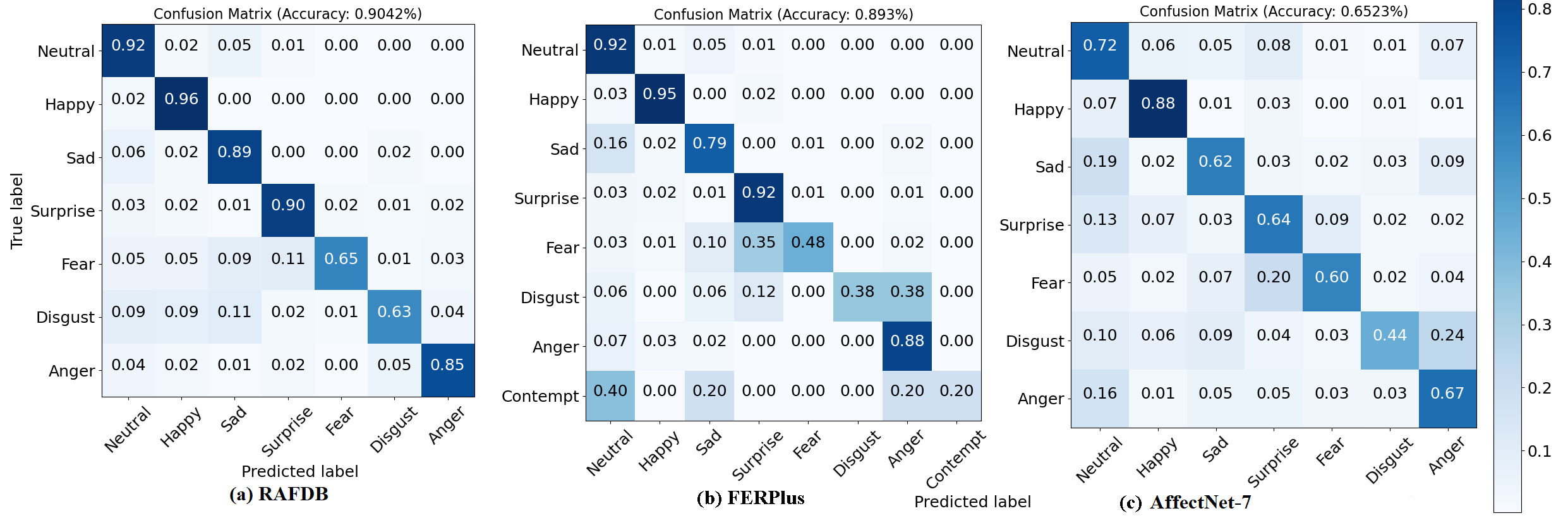}}
\caption{Confusion plots for benchmark FER datasets RAFDB, FERPlus and AffectNet-7.}
\label{confusion_plots}
\end{figure*}

\begin{figure}[hbt!]
\centerline{\includegraphics[width=3.5in]{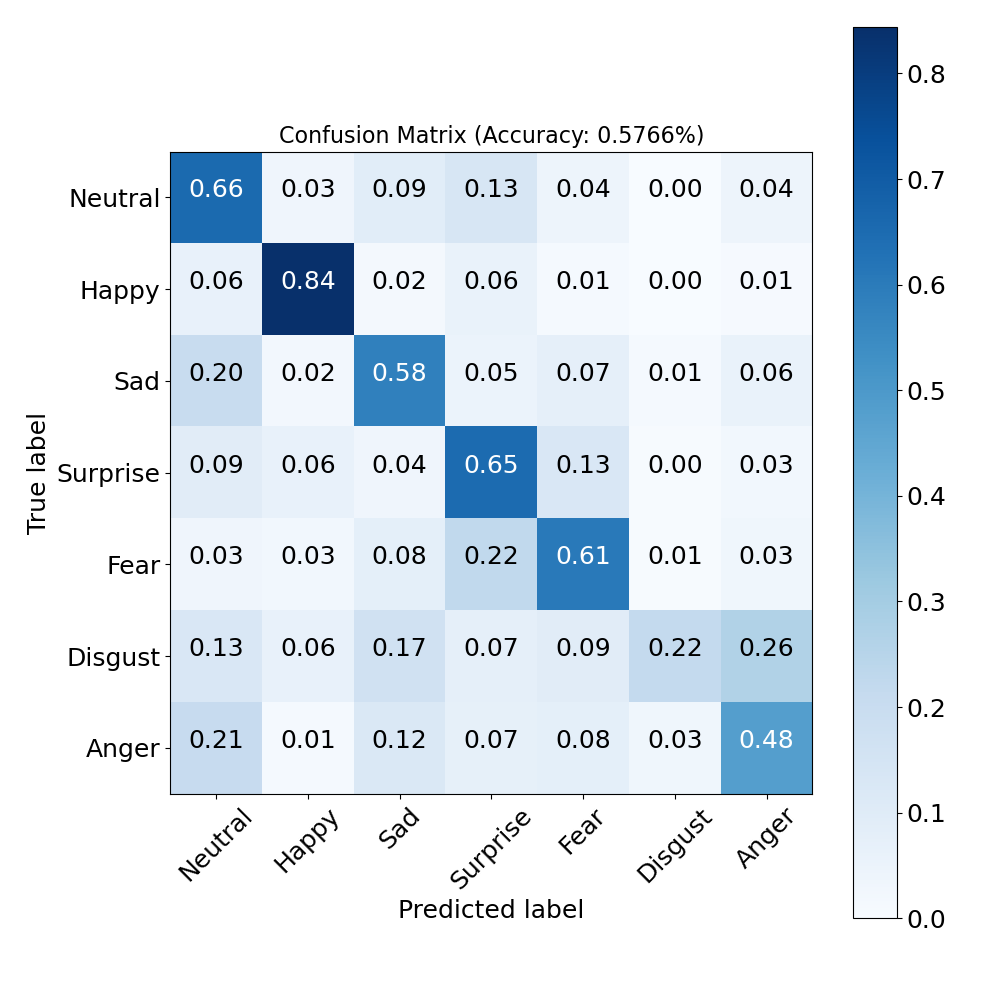}}
\caption{Confusion plots for AffectNet-7 by training on real noisy AffectNet (automatically annotated) train subset.}
\label{real_confusion_plots}
\end{figure}

In addition to the overall accuracy, we also analysed the individual expression recognition performance of DNFER on benchmark FER datasets RAFDB, FERplus and AffectNet (shown in Fig. \ref{confusion_plots}) as well as on real-noisy annotated AffectNet-7 (shown in Fig. \ref{real_confusion_plots}). Clearly, Happiness is the easiest expression to recognize across all the datasets followed by Neutral, Surprise and Anger whereas Fear and Disgust are difficult expressions. However, contempt is the most difficult expression in FERPlus, exhibiting high degree of confusion with neutral. On real noisy annotated AffectNet-7, Neutral, Happiness and Surprise are easy expressions whereas Anger and Disgust are difficult ones. Disgust is often confused with Anger whereas Anger is confused with Neutral. Overall, DNFER performs consistently in discriminating expressions in synthetic and real noisy annotated datasets as well as on clean datasets. DNFER exhibits superior or comparable performance on all the benchmark FER datasets.

\subsubsection{t-SNE plots}
\begin{figure*}[hbt!]
\centerline{\includegraphics[width=5.2in]{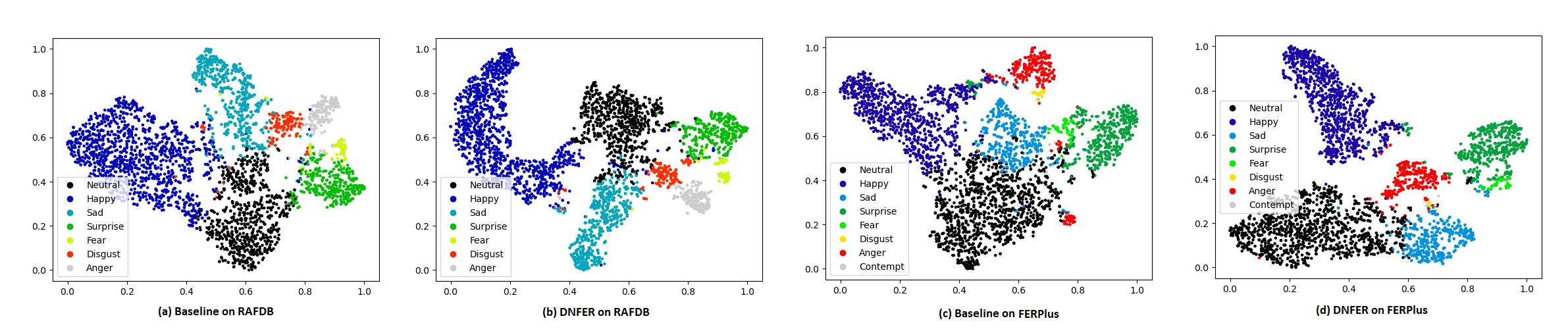}}
\caption{t-SNE \cite{tsne} visualizations of facial expression features obtained by (a) Baseline and (b) DNFER on RAFDB test set; and (c) Baseline and (d) DNFER on FERPlus test set. Models are trained on corresponding trained sets.}
\label{tsne}
\end{figure*}

In order to evaluate the effectiveness of proposed DNFER method, we use t-SNE \cite{tsne} to visualize the facial features distribution extracted in 2-D space by Baseline in Fig. \ref{tsne} (a) and (c) on RAFDB and FERPlus respectively. These are compared with corresponding feature distribution learnt using DNFER in Fig. (b) and (d) respectively. We can observe that facial expression features using Baseline are not discriminating due to large intra-class variation as well as inter-class variation. In contrast, expression features using DNFER have high intra-class similarity and inter-class differences, thus giving superior performance. 
\subsection{Comparison with the State-of-the-art}
We compare proposed DNFER method with existing SOTA methods on 4 popular in-the-wild FER benchmark datasets, RAFDB (see Table \ref{SOTA_rafdb}), FERPlus (see Table \ref{SOTA_Ferplus}), SFEW (see Table \ref{SOTA_SFEW}) and AffectNet (see Tables \ref{SOTA_AffectNet7} and \ref{SOTA_AffectNet8}) datasets. Our DNFER obtains top performance on RAFDB (90.41\%) compared to other SOTA methods (an improvement by a margin of 0.94 w.r.t top performing FDRL). Although DMUE, RUL, SCN and ULC-AG are designed to deal with uncertain labels, their performance is inferior comapred to DNFER which utilizes all the samples for feature learning. Our DNFER obtains comparable performance on FERPlus (89.32\%). On SFEW, DNFER obtains (57.8\%) same performance as that of recent top performing model LTLAttenNet. It achieves competitive performance on AffectNet-8 (61.6\%) and Affectnet-7 (65.22\%) compared to recent SOTA methods. These remarkable results demonstrate the effectiveness of proposed DNFER to deal with noisy annotations present in real-world FER datasets.

\begin{table}[hbt!]
  
     \caption{Comparison on RAFDB (* denotes FER methods with noisy labels robust learning) }
     \centering
     \begin{tabular}{|c|r|r|}
        \hline
        Method       & Year & Accuracy (\%)       \\
        \hline
        RAN \cite{ran}       & 2020  & 86.90      \\
        IPA2LT \cite{ipa2lt}       & 2020  & 86.77      \\
        THIN \cite{thin}     & 2022  & 87.81  \\
        Mahmoudi et al. \cite{kdl}     & 2020  & 88.02  \\
        Liu et. \cite{point}  & 2020  & 88.02  \\
        SCN* \cite{scn}    & 2020  & 88.14       \\
        OADN \cite{oadn}    & 2020  & 89.83       \\
        GCN \cite{gcn}      & 2020    & 89.41       \\
        SCAN \cite{scan}      & 2021    & 89.02      \\
        FER-VT \cite{fervt}  & 2021 & 88.26 \\
        DMUE* \cite{DMUE}      & 2021 & 88.76 \\
        RUL* \cite{RUL}  & 2021 & 88.98 \\
        EfficentFace \cite{EfficientFace} & 2021 & 88.36 \\
        
        FDRL \cite{FDRL} & 2021 & 89.47 \\
        ULC-AG* \cite{ulcag}              & 2022 & 89.31 \\
        \hline
         DNFER* &          & \textbf{90.41}   \\
         
        \hline
    \end{tabular}
  
     \label{SOTA_rafdb}
\end{table}
\begin{table}[hbt!]
  
     \caption{Comparison on SFEW }
     \centering
     \begin{tabular}{|c|r|r|}
        \hline
        Method       & Year & Accuracy (\%)       \\
        \hline
        IdentityCNN \cite{identitynet} & 2019 & 50.98 \\
        Island loss \cite{island_loss} & 2018 & 52.52 \\
        RAN  \cite{ran}    & 2020 & 56.40 \\
        DMUE* \cite{DMUE}& 2021 & 57.12\\
        LNLAttenNet \cite{LNLAttenNet}& 2022 & 57.8 \\
        \hline
         DNFER* &          & \textbf{57.77}   \\
         
        \hline
    \end{tabular}
  
     \label{SOTA_SFEW}
\end{table}

\begin{table}[hbt!]
\caption{Comparison on FERPlus (* denotes FER methods with noisy labels robust learning)}
     \centering
    \begin{tabular}{|c|r|r|}
        \hline
        Method       & Year & Accuracy (\%)       \\
        \hline
         Georgescu et al.\cite{43_georgescu2019local}  & 2019 & 87.76\\
        RAN \cite{ran}       & 2020  & 88.55      \\
        ESR \cite{esr}  & 2020 & 87.15 \\
        OADN \cite{oadn}    & 2020  & 88.71$^{*}$    \\
        SCN* \cite{scn}    & 2020  & 88.01      \\
        GCN \cite{gcn}      & 2020    & 89.39       \\
        SCAN \cite{scan}      & 2021    & 89.42      \\
        DMUE* \cite{DMUE}      & 2021 & 88.64 \\
        LNLAttenNet \cite{LNLAttenNet} & 2022 & 86.15 \\
         \hline
        DNFER       &     &   \textbf{89.32}   \\
        \hline
    \end{tabular}

     \label{SOTA_Ferplus}
\end{table}

\begin{table} [hbt!] 
\caption{Comparison on AffectNet-7}
  \centering
      \begin{tabular}{|c|r|r|r|}
        \hline
        Method       & Year & Accuracy (\%)       \\
        \hline
        Annotators Agreement \cite{affectnet} & 2017 & 65.3 \\
        \hline
         HERO \cite{hero}       & 2019 & 62.11 \\
         Yongjian et al. \cite{semantic}       & 2019  &   62.7    \\
         FMPN \cite{fmpn}       & 2019  &   61.5    \\
          Georgescu et al.\cite{43_georgescu2019local}  & 2019 & 63.31\\
         IPA2LT \cite{ipa2lt}       & 2020  & 57.31      \\
        
         THIN \cite{thin}     & 2020  & 63.97 \\
         OADN \cite{oadn}    & 2020  &  64.06       \\
         GCN \cite{gcn}      & 2020    & 64.46        \\
         SCAN \cite{scan}      & 2021    & 65.14      \\
         \hline
        DNFER        &     &   \textbf{65.22}     \\
        \hline
    \end{tabular}
 
 \label{SOTA_AffectNet7}
\end{table}

\begin{table}   [hbt!]
\caption{Comparison on AffectNet-8 (* denotes FER methods with noisy labels robust learning)}
  \centering
    \begin{tabular}{|c|r|r|r|}
        \hline
        Method       & Year & Accuracy (\%)       \\
        \hline
        Georgescu et al. \cite{43_georgescu2019local}  & 2019 & 59.58\\
        RAN \cite{ran}       & 2020  &59.50     \\
         ESR \cite{esr}  & 2020 & 59.3 \\
           SCN \cite{scn}    & 2020  & 60.23      \\
        GCN \cite{gcn}      & 2020    & 60.58      \\
        SCAN \cite{scan}      & 2021    & 61.73      \\
        ULC-AG \cite{ulcag}             & 2022 & 61.57 \\
        DMUE* \cite{DMUE}     & 2021 & 62.84 \\
        LNLAttenNet \cite{LNLAttenNet} & 2022 & 59.28 \\
         \hline
        DNFER       &   -  &    \textbf{ 61.67}  \\
        \hline
    \end{tabular}

 \label{SOTA_AffectNet8}
\end{table}

\section{Conclusion}  \label{Section_V}
In this paper, we propose a novel and effective Dynamic Adaptive Threshold (DNFER) framework to learn expression discriminating features robust to noisy annotations. DNFER selects clean samples based on dynamic class-specific threshold for supervised learning, thus mitigating the influence of noisy annotations samples to prevent overfitting as well as imbalance of classes in terms of intra-class difficulty and inter-class sample size. In addition, it uses all the samples by aligning the posterior distributions of weakly-augmented and strongly-augmented image using a consistency loss for expression feature representation. Extensive experiments on synthetic as well as real noisy annotated FER datasets demonstrate its effectiveness in handling noisy annotations. Further, SOTA performance on popular benchmark FER datasets validates the utility of DNFER as a general purpose FER method. In future, DNFER can be extended to handle the task of AU prediction and valence-arousal estimation in multi-task setting.

\section{Acknowledgments}
We dedicate this work to Our Guru and Guide Bhagawan Sri Sathya Sai Baba, Divine Founder Chancellor of Sri Sathya Sai Institute of Higher Learning, Prasanthi Nilayam, Andhra Pradesh, India. 

\bibliographystyle{unsrt}
\bibliography{references}  
\end{document}